\documentclass[10pt,twocolumn,letterpaper]{article}

\usepackage{cvpr}              % To produce the CAMERA-READY version
\definecolor{cvprblue}{rgb}{0.21,0.49,0.74}
\usepackage[pagebackref,breaklinks,colorlinks]{hyperref}
\usepackage{fontawesome5} % 图标字体
\usepackage{hyperref}
\def\paperID{*****} % *** Enter the Paper ID here
\def\confName{CVPR}
\def\confYear{2026}
\usepackage[table,xcdraw]{xcolor}
\usepackage{booktabs,tabularx,threeparttable,array,makecell}
\usepackage{tabularx}
\usepackage{graphicx}
\usepackage{pifont}
\usepackage{caption}
\newcommand{\cmark}{\ding{51}} % ✓
\newcommand{\xmark}{\ding{55}} % ✗
\newcommand{\pmark}{\textcolor{orange!80!black}{Partial}}
\usepackage{dblfloatfix} 
\usepackage{siunitx}
\usepackage{threeparttable}
\usepackage{multirow}
\usepackage{float}
\usepackage{adjustbox}
\usepackage{colortbl}
\definecolor{oursrow}{RGB}{235,229,247}
\definecolor{Rule}{HTML}{9AA4B2}
\arrayrulecolor{Rule}
\renewcommand{\arraystretch}{1.10}

\usepackage{makecell}
\usepackage{colortbl}
\definecolor{Header}{HTML}{0F172A}   % deep slate
\definecolor{Rule}{HTML}{9AA4B2}     % soft grey for rules
\renewcommand{\arraystretch}{1.10}
\rowcolors{3}{gray!2}{white}
\arrayrulecolor{Rule}
\newcolumntype{C}[1]{>{\centering\arraybackslash}p{#1}}
\newcolumntype{Y}{>{\centering\arraybackslash}X}

\definecolor{headbg}{RGB}{246,248,252}   % header background
\definecolor{sectbg}{RGB}{244,244,246}   % section band ("Proprietary", "Open-Source")
\definecolor{memcol}{RGB}{232,243,255}   % Memory block
\definecolor{undcol}{RGB}{235,251,242}   % Understanding block
\definecolor{crosscol}{RGB}{255,243,231} % Cross-Memory block
\title{\raisebox{-0.2\height}{\includegraphics[height=5em]{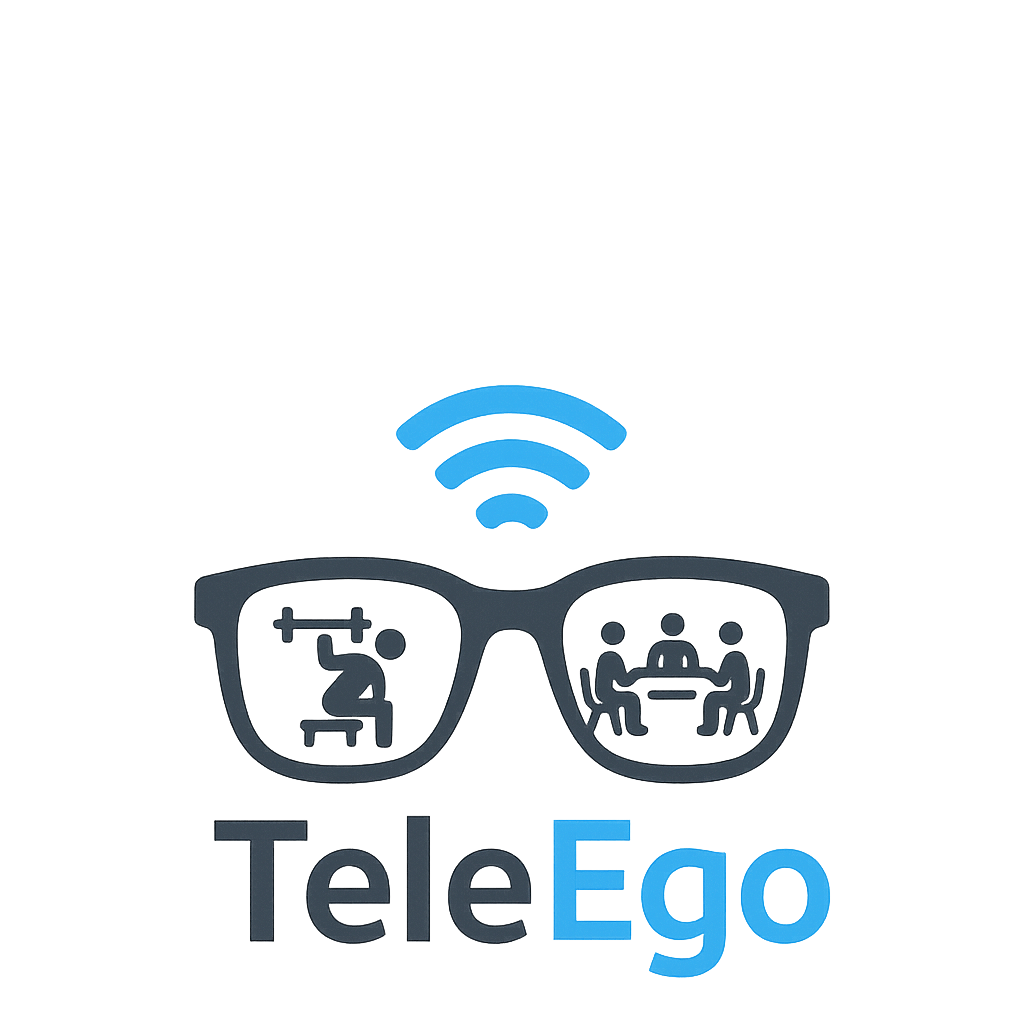}}\hspace{-0.8em}
TeleEgo: Benchmarking Egocentric AI Assistants in the Wild
\vspace{-5pt}
}

\author{
\textbf{Jiaqi Yan}$^{*}$,
\textbf{Ruilong Ren}$^{*}$, 
\textbf{Jingren Liu}$^{*}$,
\textbf{Shuning Xu},
\textbf{Ling Wang},\\
\textbf{Yiheng Wang}, 
\textbf{Xinlin Zhong},
\textbf{Yun Wang},
\textbf{Long Zhang},
\textbf{Xiangyu Chen}${^{\dagger}}$,\\
\textbf{Changzhi Sun},
\textbf{Jixiang Luo},
\textbf{Dell Zhang}$^{\#}$,
\textbf{Hao Sun},
\textbf{Chi Zhang}, 
\textbf{Xuelong Li}$^{\#}$
\\
Institute of Artificial Intelligence (TeleAI), China Telecom\\
\small\url{https://github.com/TeleAI-UAGI/TeleEgo}
}

\begin{document}
\maketitle
{\let\thefootnote\relax\footnotetext{\noindent* Equal contribution. $^{\dagger}$ Project lead. $^{\#}$ Corresponding authors.}}

\begin{abstract}
Egocentric AI assistants in real-world settings must process multi-modal inputs (video, audio, text), respond in real time, and retain evolving long-term memory. However, existing benchmarks typically evaluate these abilities in isolation, lack realistic streaming scenarios, or support only short-term tasks. 
We introduce \textbf{TeleEgo}, a long-duration, streaming, omni-modal benchmark for evaluating egocentric AI assistants in realistic daily contexts. The dataset features over 14 hours per participant of synchronized egocentric video, audio, and text across four domains: work \& study, lifestyle \& routines, social activities, and outings \& culture. All data is aligned on a unified global timeline and includes high-quality visual narrations and speech transcripts, curated through human refinement.
TeleEgo defines 12 diagnostic subtasks across three core capabilities: Memory (recalling past events), Understanding (interpreting the current moment), and Cross-Memory Reasoning (linking distant events). It contains 3,291 human-verified QA items spanning multiple question formats (single-choice, binary, multi-choice, and open-ended), evaluated strictly in a streaming setting.
%
% We propose Real-Time Accuracy to measure correctness under tight decision windows and Memory Persistence Time as a forward-looking metric of long-term retention in continuous streams. Our experiments focus on RTA, and we release TeleEgo and an MPT evaluation framework as a realistic, extensible benchmark for future egocentric AI assistants.
% We propose Real-Time Accuracy (RTA) to jointly capture correctness and responsiveness under tight decision windows, and Memory Persistence Time (MPT) as a forward-looking metric for long-term retention in continuous streams. In this work, we report RTA results for current models and release TeleEgo, together with an MPT evaluation framework, as a realistic and extensible benchmark for future egocentric assistants with stronger streaming memory, enabling systematic study of both real-time behavior and long-horizon memory.
We propose Real-Time Accuracy (RTA) to jointly capture correctness and responsiveness under tight decision windows, and Memory Persistence Time (MPT) as a forward-looking metric for long-term retention in continuous streams. In this work, we report RTA results for current models and release TeleEgo, together with an MPT evaluation framework, as a realistic and extensible benchmark for future egocentric assistants with stronger streaming memory, enabling systematic study of both real-time behavior and long-horizon memory.

\end{abstract}    
\vspace{-2ex}
\section{Introduction}
\label{sec:intro}
% As artificial intelligence continues to advance, egocentric AI assistants are transitioning from controlled experimental settings to real-world daily usage. To be truly effective in these real-life environments, such assistants must simultaneously demonstrate three tightly-coupled capability: i) \emph{growing memory} that spans hours or even days; ii) \emph{real-time decision-making} under continuous, streaming input; and iii) \emph{omni-modal understanding} that integrates what the camera sees, what the microphone hears, and what users say. Importantly, these capability are not exercised in isolation—they must work together in harmony. A model that remembers past events but responds too early may still produce incorrect answers; conversely, a model that correctly identifies the present moment but fails to associate the right speaker or object also falls short.

With the rapid advancement of artificial intelligence, egocentric AI assistants—those operating from a first-person perspective—are gradually transitioning from controlled experimental settings to real-world applications. To function effectively in such scenarios, these assistants must exhibit three tightly integrated capability: memory, streaming decision-making, and multimodal understanding. They must be able to retain and recall growing streams of past information; make timely judgments in the context of continuous audio-visual inputs; and interpret, in a unified manner, what the camera sees, what the microphone hears, and what the user expresses through language.
Importantly, these capability are not exercised in isolation—they must work together in harmony. A model that remembers past events but acts at the wrong moment can still fail. Similarly, a system that processes the current frame but cannot identify the speaker or key objects is unlikely to succeed. Effective real-world AI assistants must therefore reason not only about what is happening, but also when and how to respond, based on long-term context and multimodal cues.

\begin{figure*}[htbp]
    \centering
    \includegraphics[width=\textwidth]{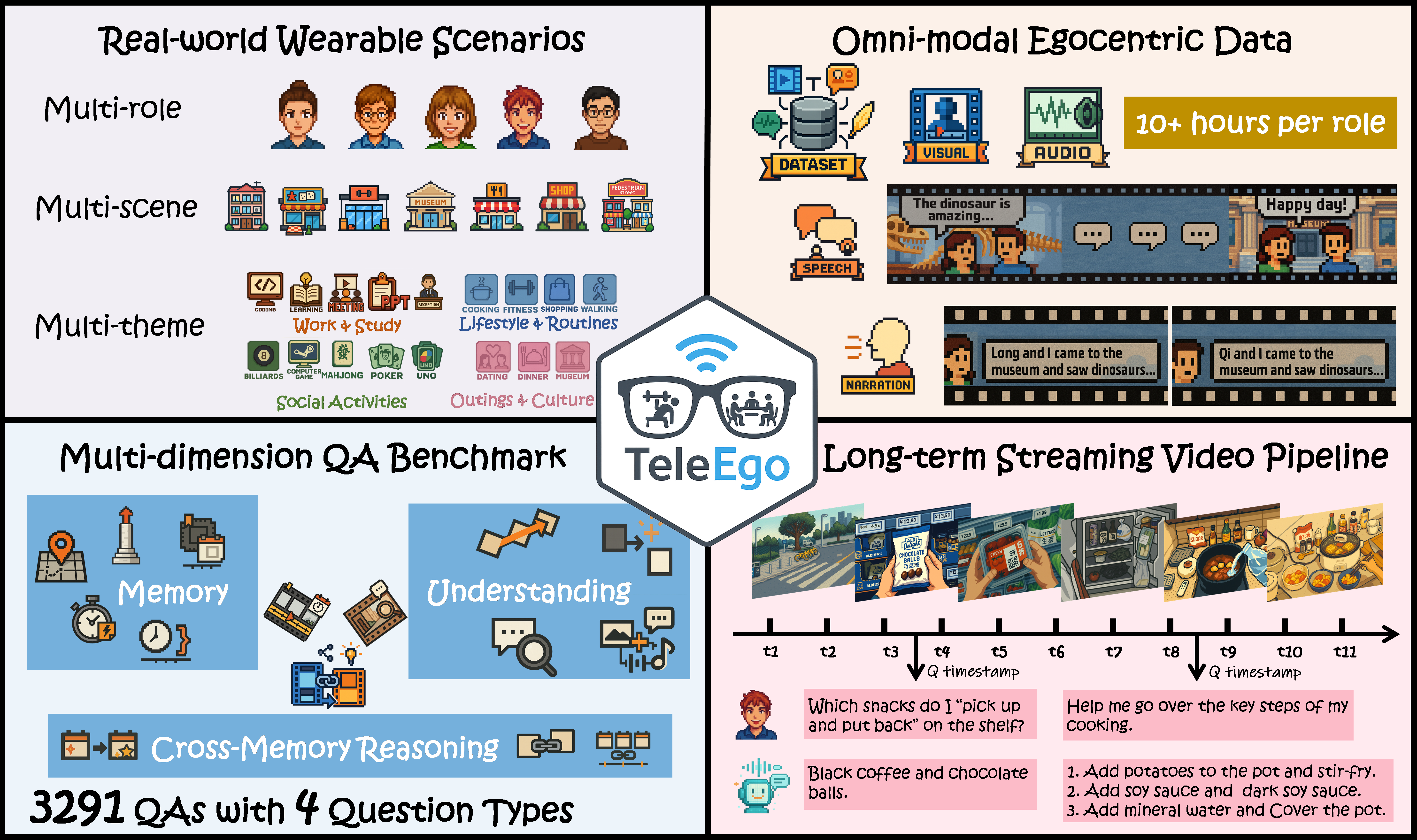}
    \caption{\textbf{An overview of the TeleEgo project.} Top-left: Scripted real-world wearable-camera scenarios covering multiple roles, scenes, and themes. Top-right: Omni-modal egocentric streaming data aligned to a shared timeline, comprising video, audio, and human-curated speech transcripts and visual narrations. Online multitask QA benchmark organized across three capability dimensions (Memory, Understanding, Cross-Memory Reasoning), containing 3,291 QA items across 4 question types. Bottom-right: Long-term streaming video pipeline—egocentric footage with query-time retrieval spanning seconds to days.}
    \vspace{-2ex}
    \label{teleego_overview}
\end{figure*}

% Despite these requirements, existing benchmarks often test these ability in isolation or in simplified settings. We identify three main limitations. First, many benchmarks suffer from offline or short-window bias—they either emphasize offline long-term memory (e.g., X-LeBench \cite{zhou2025x} ) or focus on short-term streaming performance (e.g., StreamingBench \cite{lin2024streamingbench} and VStream-QA \cite{zhang2024flash}), making joint evaluation difficult. Second, limited scenario diversity reduces generalization, as many datasets are collected in closed environments or from narrow demographic groups (e.g., EgoLife \cite{yang2025egolife}). Third, temporal fragmentation and weak supervision—through shuffled or clipped video segments and loosely verified QA tasks—undermine both realism and error analysis. Detailed benchmark comparisons are shown in Table~\ref{tab:benchmark_comp} .

Despite these requirements, existing benchmarks evaluate these ability in isolation or in simplified settings (Table~\ref{tab:benchmark_comp}). 
First, some focus on offline long-term memory (e.g., X-LeBench~\cite{zhou2025x}), while others test short-window streaming (e.g., StreamingBench~\cite{lin2024streamingbench}, VStream-QA~\cite{zhang2024flash}), making it hard to assess the trade-off between memory and real-time performance.
Second, true egocentric streaming evaluation is rare. Most datasets use third-person or static videos~\cite{huang2025online,lin2024streamingbench,niu2025ovo}, avoiding challenges like self-motion and viewpoint shifts. Some exceptions (e.g., ODV-Bench~\cite{zeng2025streamforest}) use first-person footage, but with short sequences and limited multimodality.
Third, few datasets offer long, continuous, real-world recordings. Many are short clips or image sets~\cite{zeng2025streamforest,cheng2024egothink,zhou2025egotextvqa}. While X-LeBench and EgoLife~\cite{yang2025egolife} are longer, the former stitches clips, and the latter is recorded in closed and controlled environments.

To address these challenges, we present TeleEgo: a long-duration, streaming, and fully multimodal benchmark grounded in real-world scenarios, purpose-built to evaluate egocentric AI assistants (see Figure \ref{teleego_overview}). TeleEgo consists of synchronized video, audio, and textual data collected from multiple participants, each contributing more than 14 hours of recordings. The dataset spans four major domains: work \& study, lifestyle \& routines, social activities, and outings \& culture. All data streams are precisely aligned to a unified global timeline and enriched with manually curated speech transcripts and visual narrations to ensure high quality and semantic clarity.
Building on this foundation, TeleEgo introduces 12 diagnostic tasks covering three core ability: \textit{memory} (recalling past events), \textit{understanding} (interpreting the present), and \textit{cross-memory reasoning} (connecting distant moments). These tasks include a total of 3,291 human-verified QA items across various formats, including single-choice, binary, multiple-choice, and open-ended questions. Each task is tied to a specific time point and a decision window, requiring models not only to answer correctly, but also to respond at the right time. All evaluations are conducted under streaming conditions, and a task is considered successful only if the model’s first correct answer falls within the allowed window. This setup prevents models from guessing too early or answering too late, offering a more accurate measure of real-time decision-making and responsiveness.
We further introduce two evaluation metrics: Real-Time Accuracy, which measures whether the model produces a correct response within a given decision window, and Memory Persistence Time, a forward-looking metric that tracks how long past information remains correctly retrievable in a continuous stream without replaying evidences. In this work, we primarily report RTA on existing models and provide MPT as an evaluation protocol for future systems with stronger long-term streaming memory.

\begin{table*}[htbp]
\centering
\resizebox{\textwidth}{!}{
\begin{tabular}{cccccccc}
\toprule
\rowcolor{gray!12}
\textbf{Benchmark} & \textbf{Focus Scene} &
\textbf{Video Duration} & \textbf{Tasks} &
\textbf{Omni-Modal} & \textbf{Streaming} & \textbf{Egocentric} & \textbf{Long-Mem QA} \\
\midrule

\rowcolor{gray!6}
EgoExoLearn \cite{huang2024egoexolearn} & Ego–Exo Skill Assessment & \makecell[c]{avg. 13.4 min (ego)\\ avg. 4.5 min (exo)} &
Cross-view Tasks & \xmark & \xmark & \pmark & \xmark \\

\rowcolor{white}
EgoThink \cite{cheng2024egothink} & First-Person Thinking & -- &
\makecell[c]{Object, Activity, Localization, Reasoning,\\ Forecasting and planning} & \xmark & \xmark & \cmark & \xmark \\

\rowcolor{gray!6}
OVBench \cite{huang2025online} & Online Video Understanding & 30 s-1 h &
Spatiotemporal Understanding and Interpretation & \xmark & \cmark & \xmark & \xmark \\

\rowcolor{white}
StreamingBench \cite{lin2024streamingbench} & Online Video Understanding & 3 s-24 min &
\makecell[c]{Real-Time Visual, Omni-Source \\and Contextual Understanding} & \cmark & \cmark & \xmark & \xmark \\

% \rowcolor{gray!6}
% MM-Ego \cite{ye2024mm} & Egocentric Video QA & 30 s-1 h &
% Visual Recognition and Memory & \xmark & \xmark & \cmark & \xmark \\

\rowcolor{gray!6}
EgoTextVQA \cite{zhou2025egotextvqa} & Egocentric Scene-text & avg. 101.7 s &
Identification and Reasoning & \xmark & \xmark & \cmark & \xmark \\

\rowcolor{white}
X-LeBench \cite{zhou2025x} & Extra Long Egocentric data & 23 min–16.4 h &
\makecell[c]{Temporal Localization, Summarization,\\ Counting and Ordering} & \xmark & \xmark & \cmark & \cmark \\

\rowcolor{gray!6}
VStream-QA \cite{zhang2024flash} & Online Video Stream & avg. 40 min &
Event and Scene Understanding & \xmark & \cmark & \pmark & \xmark \\

\rowcolor{white}
OVO-Bench \cite{niu2025ovo} & Online Video Reasoning & avg. 263.42s &
\makecell[c]{Backward Tracing, Real-Time Visual\\
Perception, and Forward Active Responding} & \xmark & \cmark & \xmark & \xmark \\

\rowcolor{gray!6}
ODV-Bench \cite{zeng2025streamforest} & Online Driving Video & 5-90 s & Realtime Perception and Prediction & \xmark & \cmark & \cmark & \xmark \\

\rowcolor{white}
EgoLife \cite{yang2025egolife} & Egocentric Assistant & avg. 44.3 h &
\makecell[c]{EntityLog, EventRecall, HabitInsight,\\RelationMap and TaskMaster} & \cmark & \xmark & \cmark & \cmark \\

\rowcolor{gray!6}
M3-Bench \cite{long2025seeing} & Agent Memory and Reasoning & avg. 34 min &
 Agent-centric QA & \cmark & \cmark & \pmark & \cmark \\
 
\rowcolor{white}
\textbf{TeleEgo (Ours)} & Real-world Settings & avg. 14.4 h &
 Memory, Understanding and Cross-Memory Reasoning & \cmark & \cmark & \cmark & \cmark \\
\bottomrule
\end{tabular}
}
% \caption{\textbf{Comparison of long-term egocentric and streaming video QA benchmarks.}
% TeleEgo unifies omni-modal, streaming, egocentric, and long-memory QA.
% \cmark denotes supported, \xmark \, not applicable, and \pmark \, partial support.}
\caption{\textbf{Related benchmarks for TeleEgo.} Comparison across duration, egocentricity, streaming protocol, modality coverage, and long-memory QA. TeleEgo (ours) uniquely satisfies all dimensions. 
\cmark denotes supported, \xmark \, not applicable, and \pmark \, partial support.}
\vspace{-2ex}
\label{tab:benchmark_comp}
\end{table*}

In general, our contributions are following:
\begin{itemize}
\item A unified, streaming, multimodal benchmark aligned to a global timeline with rich, real-world data across multiple participants and scenarios.
\item A diagnostic task suite covering memory, understanding, and cross-memory reasoning, with timestamped, multimodal evidence for interpretability.
\item An evaluation protocol for streaming inputs, using Real-Time Accuracy metric to assess contemporary models and a forward-looking Memory Persistence Time metric to study long-term retention.
\end{itemize}
\begin{figure*}[t]
    \centering
    \includegraphics[width=\textwidth]{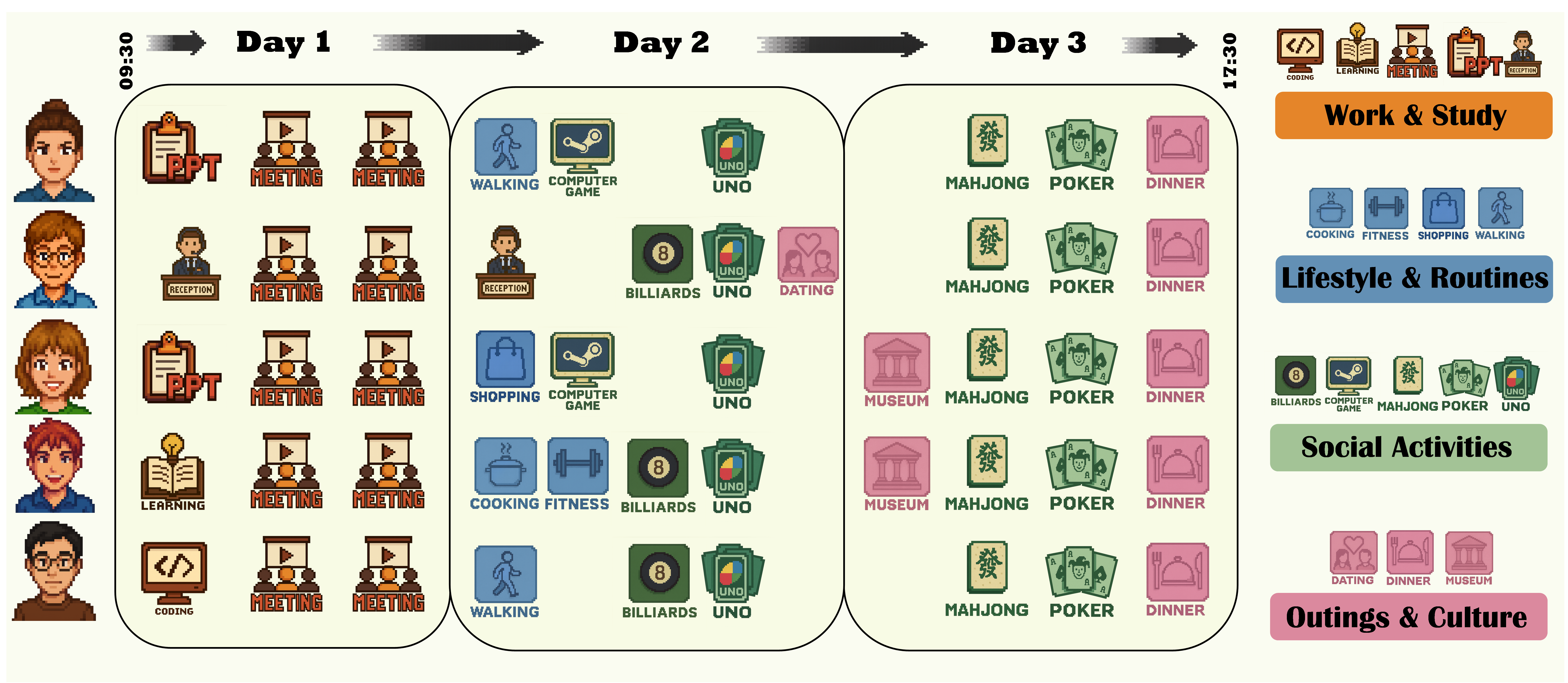}
    \caption{Scenario design and activity distribution in TeleEgo dataset. Each role engaged in diverse first-person activities across three recording days, systematically covering four themes, Work \& Study, Lifestyle \& Routines, Social Activities, and Outings \& Culture. The design spans a wide spectrum of cognitive and social contexts, combining solo and multi-role interactions across indoor and outdoor environments. This structure ensures ecological diversity and supports analyses of long-term, cross-situational understanding.}
    \vspace{-2ex}
    \label{fig:scenario}
\end{figure*}

\section{Related Works}
\label{sec:related}

\textbf{Egocentric Models and Benchmarks.}
Egocentric (first-person) vision research has grown significantly, evolving from early single-user recordings to diverse, large-scale datasets capturing daily life from a first-person perspective. Pioneering work such as EPIC-KITCHENS \cite{damen2018scaling} and its extensions like VISOR \cite{darkhalil2022epic} introduced large-scale, object-rich video datasets in home environments. Later, Ego4D and Ego-Exo4D \cite{grauman2022ego4d,grauman2024ego} broadened the scope to include tasks like episodic memory, future prediction, and skill learning, incorporating both egocentric and exocentric views.
Recent datasets have begun to explore more complex cognitive tasks and assistant-like interactions. EgoLife \cite{yang2025egolife} captures long-duration recordings from a single household, while EgoThink \cite{cheng2024egothink} and EgoExoLearn \cite{huang2024egoexolearn} focus on segment-level reasoning and teaching-following dynamics. Other work like MM-Ego \cite{ye2024mm} and EgoTextVQA \cite{zhou2025egotextvqa} targets memory and text-based understanding. However, most of these evaluations remain offline and task-specific. 
TeleEgo advances egocentric benchmarking by combining multi-day, multi-role, and multi-theme recordings with dual text annotations aligned to a unified global timeline. Its online evaluation protocol measures not only real-time decision-making, but also the persistence of memory over time—enabling more realistic evaluations of assistant capabilities.

\textbf{Streaming Video Understanding Benchmarks.}
To assess assistants in time-sensitive and dynamic environments, several benchmarks have emerged that focus on streaming video understanding. StreamingBench \cite{lin2024streamingbench}, OVBench, and OVO-Bench \cite{huang2025online,niu2025ovo} support online task formats, but typically span short episodes and lack sustained memory testing across events. ODV-Bench \cite{zeng2025streamforest} emphasizes driving tasks and short-term prediction, prioritizing perception over memory.
Some datasets, such as X-LeBench \cite{zhou2025x}, extend video QA to longer contexts, but still operate in offline settings without real-time constraints.
TeleEgo fills this gap by offering continuous, multi-day egocentric video streams with temporally grounded question-answer pairs. Its evaluation protocol emphasizes real-time responses and long-term memory recall, supporting deeper analyses of assistant performance in realistic, ever-evolving scenarios.

\textbf{Omni-Modal Assistants.}
Recent omni-modal models seek to unify text, vision, audio, and speech for more flexible assistants. Closed-source systems such as GPT-4o \cite{gpt4o} and Gemini 1.5/2.5 \cite{team2024gemini,gemini2.5pro} show strong empirical performance on speech and video, but their architectures, memory mechanisms, and streaming behavior are proprietary and largely unknown.
Open-source omni-modal models have also progressed rapidly. LLaVA--OneVision \cite{li2024llava} and InternVL-2.5 \cite{internvl-2.5} support multi-image and video understanding, while Qwen2.5--Omni \cite{xu2025qwen2} and Baichuan--Omni \cite{li2025baichuan} are optimized for audiovisual input and interactive responses. Yet they are mostly evaluated on offline, task-specific benchmarks or short clips rather than long, continuous egocentric streams.
MiniCPM--o \cite{team2025minicpm} moves closer to streaming by providing an online interface for token- and frame-level inputs, but relies on a growing KV cache without effective long-term memory control. Beyond single-model interfaces, M3-Agent \cite{long2025seeing} introduces a multimodal agent with explicit long-term memory, organizing episodic and semantic information in an entity-centric, multimodal form. However, it still assumes segment-wise video input instead of second-by-second egocentric streams.
Despite these advances, few works test whether omni-modal assistants can both respond accurately under time-sensitive decision windows and retain information over long temporal spans in realistic, continuous settings. TeleEgo combines egocentric recordings, streaming video contexts, and real-time memory-oriented evaluation to provide a unified testbed bridging egocentric datasets, streaming benchmarks, and multimodal models.

\section{TeleEgo}
\label{sec:dataset}

\subsection{Dataset Overview}
To ensure broad diversity and real-world relevance, TeleEgo uses a carefully designed data collection protocol that spans multiple roles, themes, and tasks. We recruited five participants with balanced gender representation and a wide range of cultural, regional, and personality backgrounds, aiming to reflect a representative slice of the general population.
Each participant wore a first-person camera over three consecutive days and recorded egocentric video following a set of predefined scenarios. The recordings include both solo activities and group interactions, taking place in diverse indoor and outdoor environments and across various social contexts. This approach goes beyond traditional single-household or uniform-group datasets, improving the generalizability and practical value of the benchmark.
To capture the richness of everyday human experience, we structured the data around four common life themes (see Figure~\ref{fig:scenario}). These themes cover different cognitive demands, social situations, environmental settings, and physical activities.

\noindent\textbf{Work \& Study} (e.g., giving a presentation, meetings, coding, learning, reception tasks): This category includes knowledge-based and goal-driven tasks that often involve tools, screens, and structured interactions. These activities require focused attention, task switching, and formal turn-taking, making them ideal for evaluating cognitive workload and procedural behaviors.

\noindent\textbf{Social Activities} (e.g., playing UNO, Mahjong, Poker, video games, billiards):
These scenarios feature multi-person interactions with competitive or turn-based structures. They are rich in gesture-speech coordination and quick context changes. This category is well-suited for analyzing gaze behavior, social cues, conversational grounding, and multimodal communication.

\noindent\textbf{Lifestyle \& Routines} (e.g., shopping, exercising, walking, cooking):
This theme involves semi-structured daily activities that combine object handling and movement in dynamic yet familiar environments. It supports research on long-term activity recognition, task progression, and inferring higher-level states such as fatigue or task completion.

\noindent\textbf{Outings \& Culture} (dining out, dating, visiting museums):
These scenarios occur in complex public environments with varying lighting, noise levels, and crowd density. They also involve subtle social norms and cultural practices. This category helps evaluate model robustness to occlusion, background noise, and unfamiliar contexts, while enabling understanding of social intent and etiquette.

Over the course of three days, each participant recorded a wide range of egocentric videos covering all four themes. The result is a rich, multi-role, multi-theme, multi-day dataset that supports research on long-term memory, context carryover, and generalization across different situations—key challenges in real-world perception. 
% As shown in Figure~\ref{fig:construction}, the data goes through a privacy-preserving pipeline for cleaning, speech transcription, and visual narration, followed by detailed annotation for the TeleEgo benchmark.

\begin{figure}[!t]
    \centering    \includegraphics[width=\columnwidth]{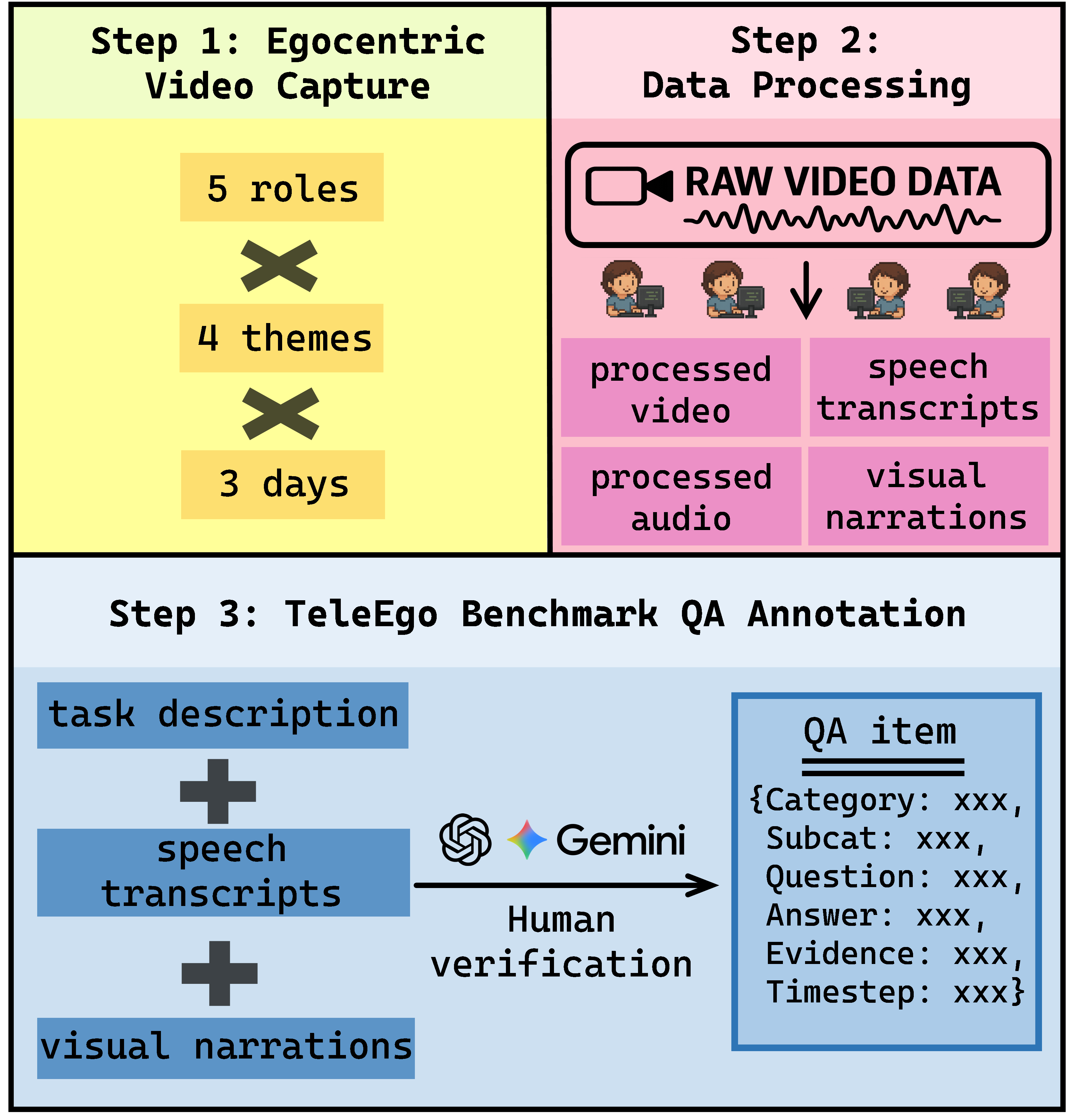}
    \caption{\textbf{TeleEgo construction pipeline.} Step 1: egocentric video capture across 5 roles, 4 themes and 3 days. Step 2: data processing into synchronized video, audio, speech, and narration captions. Step 3: AI tools generate candidate QA items from task descriptions and captions, followed by human verification.}
    \vspace{-2ex}
    \label{fig:construction}
\end{figure}

\subsection{Raw Data Processing}
To ensure privacy and ethical use, all collected recordings go through a careful de-identification process. This includes blurring faces, removing speech from non-participants, and masking any sensitive visual or audio content. These steps preserve participant privacy while keeping the recordings natural and realistic, reflecting everyday first-person experiences. To support deeper multimodal understanding beyond raw audio and video, the TeleEgo dataset includes two types of time-aligned textual annotations.

\textbf{Speech transcripts} captures all verbal communication in multi-person settings. Spoken content is automatically transcribed, then manually verified and annotated with speaker identities, yielding temporally aligned conversational transcripts that preserve discourse structure and interaction dynamics. This produces complete dialogue corpora suitable for studying social cognition, turn-taking, and multimodal grounding in egocentric contexts.

\textbf{Visual narrations} consists of participants’ self-reported verbal descriptions of their ongoing activities and salient environmental details. When explicit actions are absent, narrations focus on attentional targets and key scene elements, providing semantic coverage of visual content such as object interactions, spatial relations, and contextual cues. Each narration is timestamped and aligned with the corresponding video segment, forming a natural-language layer that parallels the perceptual stream.

Both streams are precisely synchronized with the video timeline, producing dual-layer annotations for visual and linguistic events. This structure enables rich cross-modal grounding. The processed multimodal data is then used by powerful AI tools to generate candidate QA items, which are further refined through human verification to build our benchmark system (see Figure~\ref{fig:construction}). Collectively, TeleEgo offers real-world recordings that combine perception, language, and memory, furnishing comprehensive multimodal material for evaluating AI systems’ capacity to understand and retain complex first-person experiences.

\begin{figure}[!t]
    \centering    \includegraphics[width=\columnwidth]{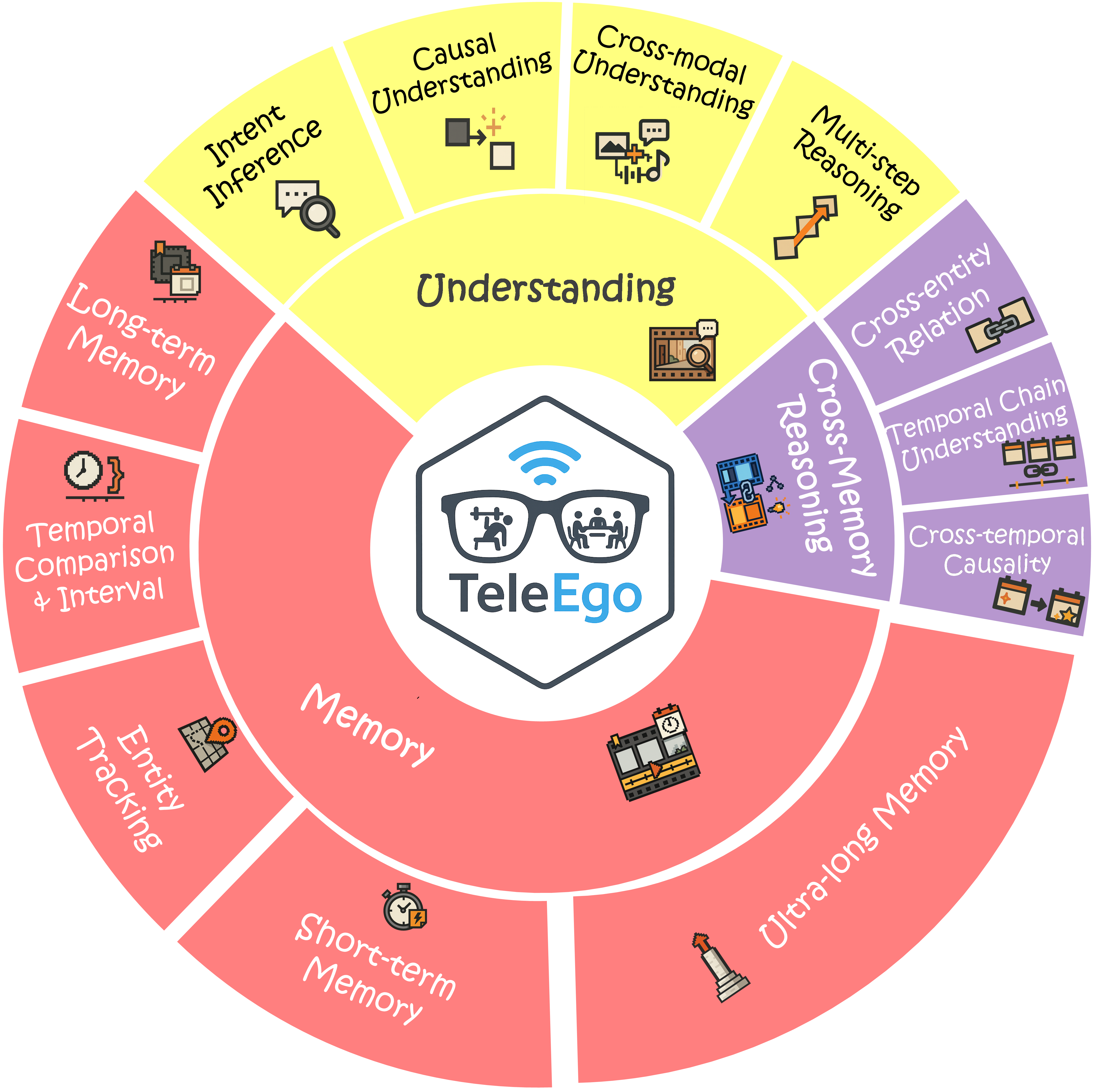}
    \caption{Hierarchical organization of the TeleEgo benchmark. The benchmark is organized around three cognitive dimensions: Memory, Understanding, and Cross-Memory Reasoning. Each dimension is further divided into fine-grained subcategories.}
    \label{fig:statistics}
    \vspace{-2ex}
\end{figure}

\subsection{Benchmark Task Design}
\begin{figure*}[!t]
    \centering
    \includegraphics[width=\textwidth]{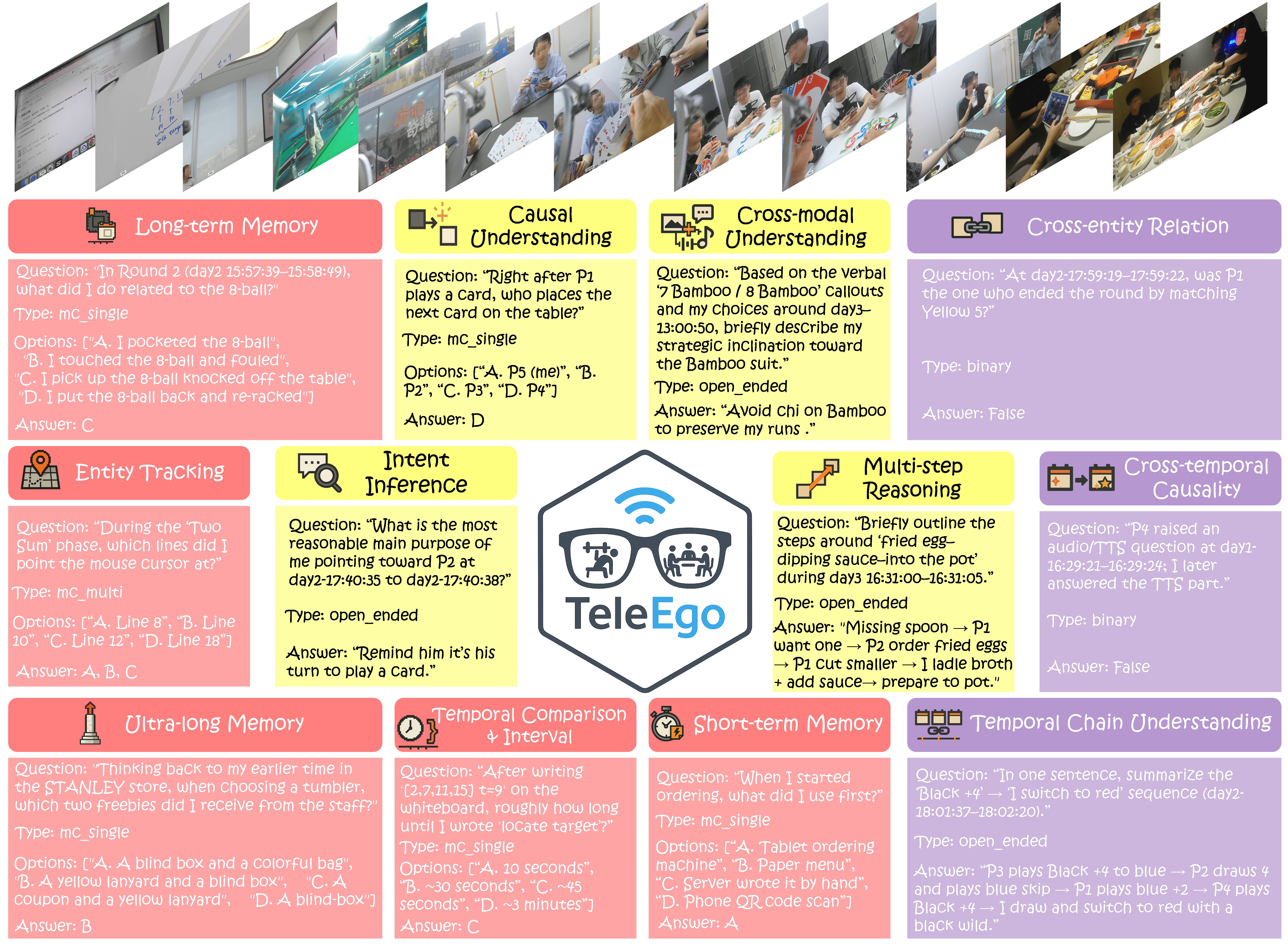}
    \caption{TeleEgo benchmark comprises twelve QA subcategories, illustrated here with one example per subcategory.}
    \label{fig:subtask}
\end{figure*}

To evaluate multidimensional cognitive abilities in egocentric video understanding, we introduce a benchmark that spans three key cognitive dimensions: memory, comprehension, and cross-memory reasoning (see Figure~\ref{fig:statistics}). These dimensions form a hierarchical structure that reflects different levels of cognition, ranging from momentary perception to long-term reasoning. This framework enables a systematic distinction between information retention, semantic understanding, and integrative reasoning across time and entities. To support this evaluation, we design 12 fine-grained question-answering subtasks, each corresponding to one of the three dimensions, allowing us to assess model performance across a wide range of cognitive scenarios. Examples of these subtasks are illustrated in Figure~\ref{fig:subtask}.

The \textbf{Memory} focuses on temporally grounded recall, assessing a model’s ability to retain, retrieve, and compare events over different time spans. Tasks range from short-term recall of transient object states and actions, to long-term and ultra-long memory over extended episodes, as well as continuous entity tracking and temporal interval reasoning. These tasks evaluate how well models maintain temporal coherence and represent evolving dynamics in first-person experiences.

\begin{table*}[!t]
  \centering
  \small % <- bigger than \footnotesize
  \begin{threeparttable}

  % ===== LEFT COLUMN (修正后) =====
  \begin{minipage}[t]{0.48\textwidth}
    {\large \bfseries Overview Panels}\par\medskip
    \centering
    % --- Category ---
    \textbf{QA Category}\par\vspace{2pt}
    % --- 改动2：移除了前后的 @{} ---
    \begin{tabularx}{\linewidth}{
      >{\centering\arraybackslash\bfseries}X
      >{\centering\arraybackslash\bfseries}S[table-format=4.0]
      >{\centering\arraybackslash\bfseries}S[table-format=2.1]}
      \toprule
      \rowcolor{gray!20}
      \multicolumn{1}{c}{\bfseries Type} & {\bfseries Count} & {\bfseries Share (\%)} \\
      \midrule
      \rowcolor{red!20}
      Memory & 1937 & 58.8 \\
      \rowcolor{yellow!20}
      Understanding & 897 & 27.3 \\
      \rowcolor{blue!10}
      Cross-Memory Reasoning & 457 & 13.9 \\
      \bottomrule
    \end{tabularx}

    \medskip
    % --- QA Types ---
    \centering
    \textbf{QA Type}\par\vspace{2pt}
    % --- 改动2：移除了前后的 @{} ---
    \begin{tabularx}{\linewidth}{
      >{\centering\arraybackslash\bfseries}X
      >{\centering\arraybackslash\bfseries}S[table-format=4.0]
      >{\centering\arraybackslash\bfseries}S[table-format=2.1]}
      \toprule
      \rowcolor{gray!20}
      \multicolumn{1}{c}{\bfseries Type} & {\bfseries Count} & {\bfseries Share (\%)} \\
      \midrule
      \rowcolor{pink!10}
      mc\_single   & 1743 & 53.0 \\
      \rowcolor{teal!10}
      binary       & 614  & 18.6 \\
      \rowcolor{orange!10}
      mc\_multi    & 493  & 15.0 \\
      \rowcolor{olive!10}
      open\_ended  & 441  & 13.4 \\
      \bottomrule
    \end{tabularx}
  \end{minipage}
  \hfill
  % ===== RIGHT COLUMN (修正后) =====
  \begin{minipage}[t]{0.48\textwidth}
    \centering
    \textbf{Subcategory Distribution}\par\vspace{2pt}
    % --- 改动2：移除了前后的 @{} ---
    \begin{tabularx}{\linewidth}{
      >{\centering\arraybackslash\bfseries}X
      >{\centering\arraybackslash\bfseries}S[table-format=4.0]
      >{\centering\arraybackslash\bfseries}S[table-format=2.1]}
      \toprule
      \rowcolor{gray!20}
      \multicolumn{1}{c}{\bfseries Subcategory} & {\bfseries Count} & {\bfseries Share (\%)} \\
      \midrule
      \rowcolor{red!20}
      Ultra-long Memory                 & 722 & 21.9 \\
      \rowcolor{red!20}
      Short-term Memory                 & 414 & 12.6 \\
      \rowcolor{red!20}
      Entity Tracking                   & 289 & 8.8 \\
      \rowcolor{red!20}
      Temporal Comparison \& Interval   & 259 & 7.9 \\
      \rowcolor{red!20}
      Long-term Memory                  & 253 & 7.7 \\
      \rowcolor{yellow!20}
      Intent Inference                  & 238 & 7.2 \\
      \rowcolor{yellow!20}
      Causal Understanding              & 225 & 6.9 \\
      \rowcolor{yellow!20}
      Cross-modal Understanding         & 219 & 6.8 \\
      \rowcolor{yellow!20}
      Multi-step Reasoning              & 215 & 6.5 \\
      \rowcolor{blue!10}
      Cross-entity Relation             & 159 & 4.8 \\
      \rowcolor{blue!10}
      Temporal Chain Understanding      & 152 & 4.6 \\
      \rowcolor{blue!10}
      Cross-temporal Causality          & 146 & 4.4 \\
      \bottomrule
    \end{tabularx}
  \end{minipage}

  \end{threeparttable}
    \caption{Statistics of TeleEgo benchmark. Left: Overview of task categories and QA types, showing a balanced mixture across Memory, Understanding, and Cross-Memory Reasoning dimensions. Right: Subcategory-level distribution over twelve cognitive tasks, encompassing a total of 3,291 QA instances. The benchmark spans short- to ultra-long memory, causal and intent reasoning, and cross-temporal integration, providing a comprehensive foundation for evaluating multimodal and embodied intelligence models.}
    % \vspace{-3ex}
    \label{tab:benchmark-unified}
\end{table*}

The \textbf{Understanding} dimension measures a model’s capacity to grasp meaning and coherence within complex, context-rich scenarios. It goes beyond surface perception to assess how well a model understands causal structures and human intent. This includes recognizing cause-effect relationships, inferring latent motivations, and constructing a unified interpretation from temporally or spatially dispersed cues. It also requires integrating multimodal inputs into coherent semantic representations.

The \textbf{Cross-Memory Reasoning} dimension challenges models to combine information across disjoint time periods and entity contexts. Tasks require building global narrative structures, linking distant events into causal chains, inferring relational dynamics between interacting agents, and synthesizing long temporal sequences into structured, meaningful processes. This dimension represents the most complex aspect of egocentric cognition, requiring reasoning over long-range dependencies in continuous experiences.

\subsection{Benchmark QA Annotation}
We adopt four complementary QA formats: single-choice (mc\_single), multiple-choice (mc\_multi), binary, and open-ended. Each format serves a distinct purpose. The mc\_single format allows for precise evaluation through carefully crafted distractors and unambiguous correct answers. The mc\_multi format captures complex or uncertain scenarios by permitting multiple correct options. Binary questions offer high-precision evaluation at low annotation cost. Open-ended questions encourage free-form reasoning and compositional thinking, complementing the more structured formats. Together, these formats strike a balance between standardization and expressiveness, enabling scalable evaluation while supporting fine-grained behavioral probing. Our QA generation process begins with post-processed, time-aligned transcripts of speech and narration. We use state-of-the-art large language models (GPT-5-Thinking and Gemini-2.5-Pro) to draft initial QA candidates. For the Ultra-Long Memory subcategory, the models ingest full dual-stream transcripts and generate questions grounded in evidence spanning 10–60 minutes. For the remaining eleven subcategories, we segment each recording into 30-minute windows and prompt the models to generate QA pairs with evidence evenly distributed across each window. Human annotators then verify factual alignment with the source video, correct timestamps, and remove ambiguous or low-quality items.

\begin{figure}[!t]
    \centering    \includegraphics[width=\columnwidth]{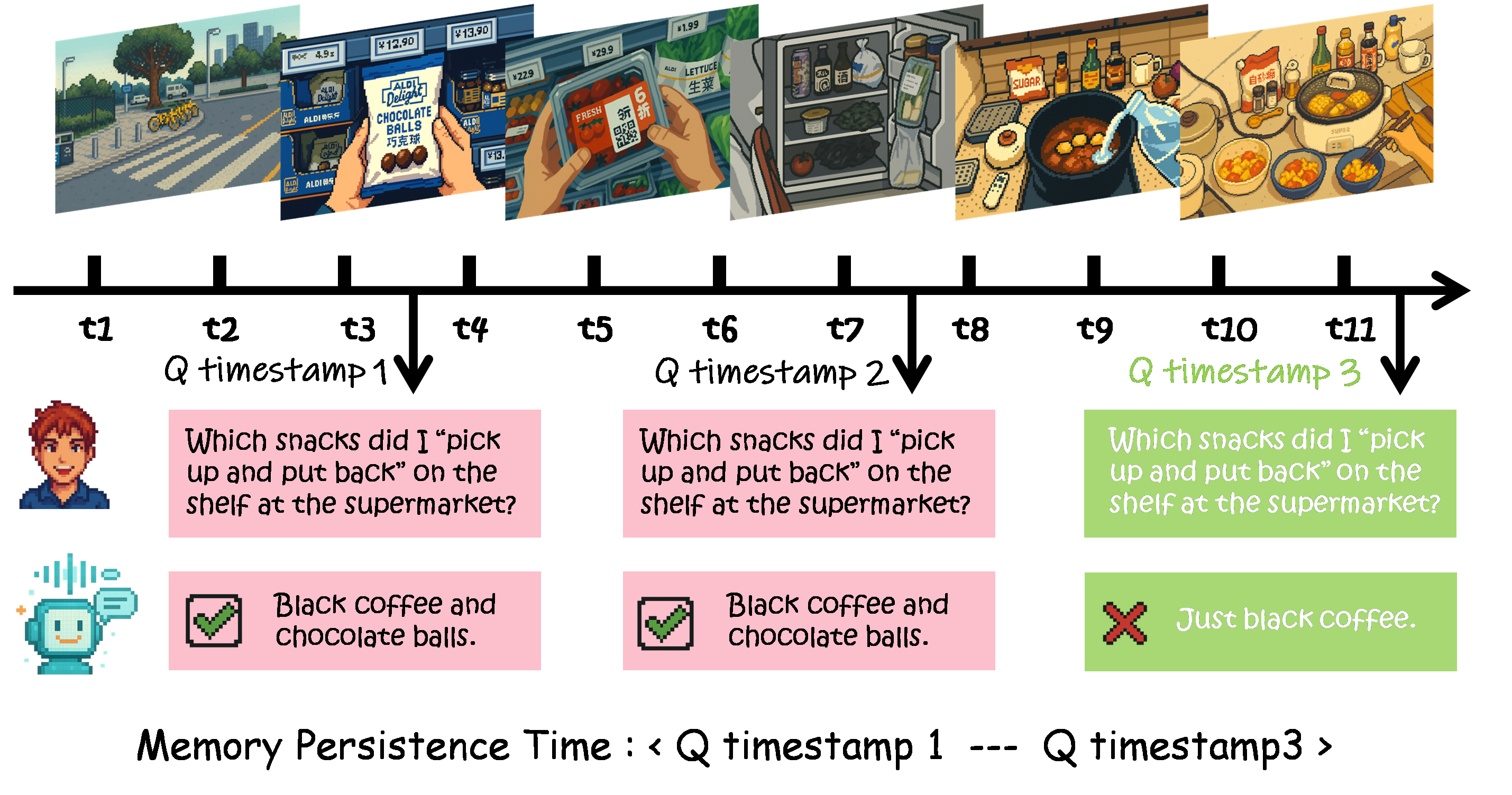}
    % \vspace{-3ex}
    \caption{Illustration of MPT test pipeline. MPT is defined as the duration from the first correct response time to the first failure.}
    \label{fig:test_pipeline}
    \vspace{-2ex}
\end{figure}

As shown in Figure~\ref{fig:statistics}, the TeleEgo benchmark is hierarchically structured along three key cognitive dimensions—memory, understanding, and reasoning—with twelve subcategories providing finer granularity. The final dataset contains 3,291 verified QA instances across all four formats (see Table~\ref{tab:benchmark-unified}). Subcategory distributions demonstrate balanced coverage across temporal, causal, and semantic reasoning challenges. Collectively, these design choices make TeleEgo a robust and discriminative benchmark for evaluating video understanding models.

\begin{table*}[!t]
\centering
\scriptsize
\setlength{\tabcolsep}{3pt}
\renewcommand{\arraystretch}{1.12}

\begin{adjustbox}{max width=\textwidth}
\rowcolors{1}{}{}
\begin{tabular}{c c c c | *{6}{c} | *{5}{c} | *{4}{c} | c}
\toprule
\rowcolor{headbg}
\multirow{2}{*}{\textbf{Method}} &
\multirow{2}{*}{\textbf{Params}} &
\multirow{2}{*}{\textbf{Omni}} &
\multirow{2}{*}{\textbf{Streaming}} &
\multicolumn{6}{c|}{\cellcolor{memcol}\textbf{Memory (\%)}} &
\multicolumn{5}{c|}{\cellcolor{undcol}\textbf{Understanding (\%)}} &
\multicolumn{4}{c|}{\cellcolor{crosscol}\textbf{Cross\textendash Memory Reasoning (\%)}} &
\multirow{2}{*}{\textbf{Overall}} \\
\cmidrule(lr){5-10}\cmidrule(lr){11-15}\cmidrule(lr){16-19}
\rowcolor{headbg}
 &  &  &  &
\cellcolor{memcol}\textbf{UlM} & \cellcolor{memcol}\textbf{StM} & \cellcolor{memcol}\textbf{ET} & \cellcolor{memcol}\textbf{TCI} & \cellcolor{memcol}\textbf{LtM} & \cellcolor{memcol}\textbf{All} &
\cellcolor{undcol}\textbf{II} & \cellcolor{undcol}\textbf{CU} & \cellcolor{undcol}\textbf{CmU} & \cellcolor{undcol}\textbf{MsR} & \cellcolor{undcol}\textbf{All} &
\cellcolor{crosscol}\textbf{CeR} & \cellcolor{crosscol}\textbf{TCU} & \cellcolor{crosscol}\textbf{CtC} & \cellcolor{crosscol}\textbf{All} &  \\
\midrule
GPT-4o \cite{gpt4o}        & -- & \cmark & -- & 42.31 & 40.58 & 31.91 & 47.37 & 52.78 & 42.18 &
81.81 & 81.58 & 45.71 & 50.00 & 66.67 &
44.44 & 40.00 & 45.00 & 44.23 &
48.94 \\
Gemini-2.5-Pro \cite{gemini2.5pro} & -- & \cmark & -- & 

49.04 & 45.59 & 34.04 & 47.37 & 44.44 & 45.05 &
63.64 & 55.26 & 37.14 & 28.57 & 48.03 &
40.74 & 40.00 & 45.00 & 42.31 &
45.55 \\
\bottomrule
\end{tabular}
\end{adjustbox}

\caption{\textbf{RTA results of proprietary MLLMs on TeleEgo (single-video evaluation).}
Columns are grouped into three capability blocks: Memory, Understanding, and Cross\textendash Memory Reasoning, with an All column summarizing each block and an Overall column aggregating across blocks. ``Omni'' denotes integrated audio--video--text perception; ``Streaming'' denotes native support for streaming interaction. A \cmark\ indicates the capability is supported, \xmark\ indicates not supported, and ``--'' indicates unknown. Due to the extensive video duration ($\sim$14 hours) and API latency, both models were evaluated on the same participant's video for fair comparison.}

% \caption{\textbf{RTA results of proprietary MLLMs on TeleEgo benchmark (P2 only).}
% Columns are grouped into three capability blocks, Memory, Understanding, and Cross\textendash Memory Reasoning, with an All column summarizing each block and an Overall column aggregating across blocks. “Omni” denotes integrated audio–video–text perception; “Streaming” denotes native support for streaming interaction. A \cmark\ indicates the capability is supported, and \xmark\ indicates not supported. Due to the long video duration ($\sim$14 hours per participant) and API latency, GPT-4o and Gemini-2.5-Pro were evaluated on P2 only. }
\label{tab:subtask-main}
\end{table*}

\vspace{1ex}

\begin{table*}[!t]
\centering
\scriptsize
\setlength{\tabcolsep}{3pt}
\renewcommand{\arraystretch}{1.12}

\begin{adjustbox}{max width=\textwidth}
\rowcolors{1}{}{}
\begin{tabular}{c c c c | *{6}{c} | *{5}{c} | *{4}{c} | c}
\toprule
\rowcolor{headbg}
\multirow{2}{*}{\textbf{Method}} &
\multirow{2}{*}{\textbf{Params}} &
\multirow{2}{*}{\textbf{Omni}} &
\multirow{2}{*}{\textbf{Streaming}} &
\multicolumn{6}{c|}{\cellcolor{memcol}\textbf{Memory (\%)}} &
\multicolumn{5}{c|}{\cellcolor{undcol}\textbf{Understanding (\%)}} &
\multicolumn{4}{c|}{\cellcolor{crosscol}\textbf{Cross\textendash Memory Reasoning (\%)}} &
\multirow{2}{*}{\textbf{Overall}} \\
\cmidrule(lr){5-10}\cmidrule(lr){11-15}\cmidrule(lr){16-19}
\rowcolor{headbg}
 &  &  &  &
\cellcolor{memcol}\textbf{UlM} & \cellcolor{memcol}\textbf{StM} & \cellcolor{memcol}\textbf{ET} & \cellcolor{memcol}\textbf{TCI} & \cellcolor{memcol}\textbf{LtM} & \cellcolor{memcol}\textbf{All} &
\cellcolor{undcol}\textbf{II} & \cellcolor{undcol}\textbf{CU} & \cellcolor{undcol}\textbf{CmU} & \cellcolor{undcol}\textbf{MsR} & \cellcolor{undcol}\textbf{All} &
\cellcolor{crosscol}\textbf{CeR} & \cellcolor{crosscol}\textbf{TCU} & \cellcolor{crosscol}\textbf{CtC} & \cellcolor{crosscol}\textbf{All} &  \\
\midrule

VideoChat-Online \cite{huang2025online} & 4B & \xmark & \multirow{3}{*}{\xmark} & 
30.74 & 26.29 & 19.35 & 29.84 & 22.78 & 26.69 &
57.35 & 44.13 & 35.44 & 27.89 & 42.34 &
18.79 & 32.00 & 41.60 & 29.43 &
31.28 \\
Qwen2.5-VL \cite{Qwen2.5-VL}& 8B & \xmark &  &
45.42 & 42.01 & 31.18 & 37.90 & 33.33 & 39.66 &
66.35 & 53.99 & 47.57 & 41.50 & 53.28 &
32.89 & 48.00 & 52.00 & 42.14 &
43.67 \\
Qwen2.5-Omni \cite{xu2025qwen2} & 7B & \cmark &  &
44.21 & 42.75 & 35.48 & 41.13 & 37.97 & 41.20 &
72.51 & 61.50 & 50.00 & 48.98 & 59.07 &
37.58 & 56.00 & 61.60 & 49.16 &
46.96 \\
\bottomrule
\end{tabular}
\end{adjustbox}

\caption{\textbf{RTA results of pseudo-streaming open-source MLLMs on TeleEgo.}
These models lack cross-call memory mechanisms and process each input unit independently. They are categorized as ``pseudo-streaming'' because they lack native support for streaming interaction. To test the models' intrinsic memory capabilities, they are evaluated under a strict streaming protocol where only the current frame and audio are provided at each second.}
\vspace{-2ex}
\label{tab:subtask-main2}
\end{table*}

\vspace{1ex}

\begin{table*}[!t]
\centering
\scriptsize
\setlength{\tabcolsep}{3pt}
\renewcommand{\arraystretch}{1.12}

% \begin{adjustbox}{max width=\textwidth}
% \rowcolors{1}{}{}
% \begin{tabular}{c c c c l| *{6}{c} | *{5}{c} | *{4}{c} | c}
% \toprule
% \rowcolor{headbg}
%  &  &  &  &  &
% \multirow{2}{*}{\textbf{Method}} &
% \multirow{2}{*}{\textbf{Params}} &
% \multirow{2}{*}{\textbf{Omni}} &
% \multirow{2}{*}{\textbf{Streaming}} &
% \multirow{2}{*}{\shortstack{\textbf{Session} \\ \textbf{Length}}} &
% \multicolumn{6}{c|}{\cellcolor{memcol}\textbf{Memory (\%)}} &
% \multicolumn{5}{c|}{\cellcolor{undcol}\textbf{Understanding (\%)}} &
% \multicolumn{4}{c|}{\cellcolor{crosscol}\textbf{Cross\textendash Memory Reasoning (\%)}} &
% \multirow{2}{*}{\textbf{Overall}} \\
% \cmidrule(lr){6-11}\cmidrule(lr){12-16}\cmidrule(lr){17-20}
% \rowcolor{headbg}
%  &  &  &  &  &
% \cellcolor{memcol}\textbf{UlM} & \cellcolor{memcol}\textbf{StM} & \cellcolor{memcol}\textbf{ET} & \cellcolor{memcol}\textbf{TCI} & \cellcolor{memcol}\textbf{LtM} & \cellcolor{memcol}\textbf{All} &
% \cellcolor{undcol}\textbf{II} & \cellcolor{undcol}\textbf{CU} & \cellcolor{undcol}\textbf{CmU} & \cellcolor{undcol}\textbf{MsR} & \cellcolor{undcol}\textbf{All} &
% \cellcolor{crosscol}\textbf{CeR} & \cellcolor{crosscol}\textbf{TCU} & \cellcolor{crosscol}\textbf{CtC} & \cellcolor{crosscol}\textbf{All} &  \\
% \midrule
\begin{adjustbox}{max width=\textwidth}
\rowcolors{1}{}{}
\begin{tabular}{c c c c c| *{6}{c} | *{5}{c} | *{4}{c} | c}
\toprule
\rowcolor{headbg}
 &  &  &  &  &
\multicolumn{6}{c|}{\cellcolor{memcol}\textbf{Memory (\%)}} &
\multicolumn{5}{c|}{\cellcolor{undcol}\textbf{Understanding (\%)}} &
\multicolumn{4}{c|}{\cellcolor{crosscol}\textbf{Cross\textendash Memory Reasoning (\%)}} &
 \\
\cmidrule(lr){6-11}\cmidrule(lr){12-16}\cmidrule(lr){17-20}
\rowcolor{headbg}
\multirow{-2}{*}{\textbf{Method}} &
\multirow{-2}{*}{\textbf{Params}} &
\multirow{-2}{*}{\textbf{Omni}} &
\multirow{-2}{*}{\textbf{Streaming}} &
\multirow{-2}{*}{\shortstack{\textbf{Session} \\ \textbf{Length}}} &
\cellcolor{memcol}\textbf{UlM} & \cellcolor{memcol}\textbf{StM} & \cellcolor{memcol}\textbf{ET} & \cellcolor{memcol}\textbf{TCI} & \cellcolor{memcol}\textbf{LtM} & \cellcolor{memcol}\textbf{All} &
\cellcolor{undcol}\textbf{II} & \cellcolor{undcol}\textbf{CU} & \cellcolor{undcol}\textbf{CmU} & \cellcolor{undcol}\textbf{MsR} & \cellcolor{undcol}\textbf{All} &
\cellcolor{crosscol}\textbf{CeR} & \cellcolor{crosscol}\textbf{TCU} & \cellcolor{crosscol}\textbf{CtC} & \cellcolor{crosscol}\textbf{All} &
\multirow{-2}{*}{\textbf{Overall}} \\
\midrule
% \multirow{6}{*}{MiniCPM-o \cite{team2025minicpm}} & \multirow{6}{*}{8B} & \multirow{6}{*}{\cmark} & \multirow{6}{*}{\cmark} &  1sec &
% 30.74 & 26.29 & 19.35 & 29.84 & 22.78 & 26.69 &
% 57.35 & 44.13 & 35.44 & 27.89 & 42.34 &
% 18.79 & 32.00 & 41.60 & 29.43 &
% 31.28 \\
& & & &  1sec &
51.99 & 51.60 & 35.13 & 47.58 & 44.30 & 47.54 &
75.83 & 61.50 & 51.94 & 44.22 & 59.59 &
28.19 & 48.00 & 64.80 & 45.15 &
50.60 \\
% & & & & 30sec &
% 51.64 & 52.09 & 36.2 & 54.84 & 48.10 & 49.26 &
% 81.04 & 69.48 & 59.71 & 54.42 & 67.18 &
% 36.24 & 40.00 & 65.60 & 48.83 &
% 54.14 \\
& & & & 1min &
50.60 & 55.53 & 37.28 & 53.63 & 51.05 & 50.11 &
80.57 & 69.48 & 56.80 & 51.02 & 65.64 &
33.56 & 36.00 & 66.40 & 47.49 &
54.10 \\
& & & & 3min &
48.19 & 52.58 & 37.28 & 49.19 & 45.99 & 47.31 &
73.93 & 63.38 & 53.88 & 40.14 & 59.33 &
32.89 & 32.00 & 66.40 & 46.82 &
50.57 \\
& & & & 5min &
43.00 & 44.23 & 33.69 & 47.58 & 37.55 & 41.71 &
63.51 & 56.34 & 44.17 & 43.54 & 52.64 &
23.49 & 32.00 & 56.80 & 38.13 &
44.34 \\
\multirow{-6}{*}{MiniCPM-o \cite{team2025minicpm}} & \multirow{-6}{*}{8B} & \multirow{-6}{*}{\cmark} & \multirow{-6}{*}{\cmark} & 10min &
28.50 & 28.01 & 23.66 & 30.65 & 26.16 & 27.60 &
41.71 & 40.38 & 34.95 & 29.93 & 37.32 &
15.44 & 36.00 & 37.60 & 26.42 &
30.15 \\
\bottomrule
\end{tabular}
\end{adjustbox}

\caption{\textbf{RTA results of MiniCPM-o with varying session lengths on TeleEgo.}
Session length determines how long the model retains memory of previously observed frames before reset. ``1s" corresponds to no memory (each call is independent). Longer sessions allow more historical context but may cause attention dilution and generation quality degradation.}
\vspace{-2ex}
\label{tab:subtask-main3}
\end{table*}

\section{Experiments}
\label{sec:exp}

\vspace{0.5ex}
\subsection{Evaluation Protocol}
\paragraph{Online-only regime.}
All evaluations adhere to the streaming protocol. Video frames and audio are timestamped to a single global clock. For question-answer (QA) instance \(i\), we define a decision window \([t^{(i)}_{\mathrm{s}},\, t^{(i)}_{\mathrm{s}}+T]\), where \(T\) denotes the time margin. At test time, the model must respond within this window. We set \(T=5\,\mathrm{s}\) in all experiments. We report two primary metrics:

\noindent\textbf{1) Real-Time Accuracy (RTA, \%)}: The percentage of QA items for which the model outputs a correct answer within the decision window.

\noindent\textbf{2) Memory Persistence Time (MPT, minutes)}: For each item correctly answered at time $t^\star$, we continue streaming without repeating the original evidence. At regular intervals, we re-query the same item. MPT is the time from $t^\star$ until the first failed recall. If the item is never answered correctly, MPT$\,{=}\,0$. If it is never forgotten within the probing range, it is right-censored at the last probe. Figure~\ref{fig:test_pipeline} provides a schematic overview of the pipeline used to compute Memory Persistence Time (MPT).

\vspace{-1.5ex}
\paragraph{Systems and settings.}
\label{sec:systems}
We evaluate six state-of-the-art vision-language or omni-modal models, covering both proprietary and open-source systems, as well as streaming-specialized designs:
Gemini-2.5-Pro \cite{gemini2.5pro}, GPT-4o \cite{gpt4o}, Qwen2.5-Omni \cite{xu2025qwen2}, Videochat-Online \cite{huang2025online}, Qwen2.5-VL \cite{Qwen2.5-VL}, and MiniCPM-o \cite{team2025minicpm}. All models receive synchronized video, audio, and text inputs unless stated otherwise. For models lacking built-in speech recognition, we attach a Whisper-style ASR component \cite{radford2023robust} to transcribe audio.

\vspace{0.5ex}
\subsection{Implementation Details}
\vspace{0.5ex}
\label{sec:impl}
\paragraph{Setup and Streaming.}
All experiments are conducted on a single NVIDIA H200 GPU (140 GB). At inference time, for each role we order all videos by their start timestamps and concatenate them into one continuous stream, mirroring realistic personal-assistant usage where the assistant is invoked intermittently around task- or context-specific segments rather than running continuously.

% \paragraph{Streaming harness.}
% Video is sampled at \textbf{1\,FPS} with adaptive densification near scene boundaries. 
% Audio is recorded at \textbf{16\,kHz} and processed into 20--40\,ms windows (10\,ms stride) over a 5--10\,s buffer. 
% In the merged-video setting, a timeline index maps ``D$k$--HH:MM(:SS)'' labels to offsets for temporal alignment.

\vspace{-2.5ex}
\paragraph{Scheduling and alignment.}
For each QA item, we take its evidence end time as the question timestamp (Qtimestamp) used for scheduling. Qtimestamps are rounded up to the nearest second. At test time, we pre-sort QA items by Qtimestamp and emit each item when the stream time first reaches its Qtimestamp; ties are resolved by source order, and emissions are confined to the item’s decision window.

\vspace{-2.5ex}
\paragraph{MPT implementation.}
For each item correctly answered at time $t^\star$, we schedule up to ten recall evaluations at $t^\star + r\Delta$ ($\Delta=60$\,s; $r=1,\dots,10$). At each evaluation, the original evidence is not replayed; only the ongoing stream is available. If an item fails an evaluation, it is removed from subsequent rounds; its horizon is the elapsed time from $t^\star$ to the first failed evaluation.

\vspace{-2.5ex}
\paragraph{RTA Evaluation Metrics.}
Multiple choice questions (mc\_single and mc\_multi) are evaluated by exact match on option letters. Binary questions are evaluated by boolean equivalence. For open-ended questions we report an LLM-judge score (0--5) produced by GPT-4o.

% \paragraph{I/O packaging.}
% Each per-second unit contains one RGB frame and the co-temporal 1\,s audio snippet (16\,kHz mono). When TTS is available, the model's spoken response is optionally streamed and saved as 16\,kHz WAV.

\subsection{Main Results}
We group our RTA analysis into three model categories, each under a distinct evaluation setting.
% We group our RTA analysis into three model categories, each under a distinct evaluation setting. as shown in Tables~\ref{tab:subtask-main}, \ref{tab:subtask-main2}, and \ref{tab:subtask-main3}.

\paragraph{Proprietary API Models.}
GPT-4o and Gemini-2.5-Pro are evaluated via API calls, with their internal implementations opaque. Due to the long video duration ($\sim$14 hours per participant) and API latency, we report RTA results on a single participant (Table~\ref{tab:subtask-main}). GPT-4o shows strong Understanding performance (66.67\%), especially on Intent Inference (81.81\%) and Causal Understanding (81.58\%). This strength aligns with the nature of Understanding tasks, which mainly require semantic reasoning and commonsense knowledge rather than precise temporal localization, allowing the model to exploit its general-purpose priors. In contrast, performance drops on Memory (42.18\%) and Cross-Memory Reasoning (44.23\%), where fine-grained temporal binding and evidence grounding are crucial.

\paragraph{Pseudo-Streaming Open-Source Models.}
VideoChat-Online, Qwen2.5-VL, and Qwen2.5-Omni lack genuine cross-call memory mechanisms (Table~\ref{tab:subtask-main2}). VideoChat-Online's Pyramid Memory Bank exists only within a single inference call and is reset at the end of each call. Qwen2.5-VL and Qwen2.5-Omni support multi-turn dialogue by naively accumulating raw conversation history, but provide no explicit streaming interface or memory management. In our evaluation, we deliberately avoid feeding uncompressed historical context into the models, since this merely shifts the memory burden to the context window rather than testing intrinsic memory capability. Instead, we adopt a strict streaming protocol: at each timestep, the model receives only the current frame and audio. Under this setting, Understanding tasks remain manageable, while Memory and Cross-Memory Reasoning become much harder. Among the three, Qwen2.5-Omni achieves the highest overall RTA (46.96\%), primarily because it has stronger base model and avoids the ASR transcription errors.

\paragraph{Streaming Model: MiniCPM-o.}
MiniCPM-o is the only model in our study with an explicit streaming interface that accepts chunk-wise incremental input (Table~\ref{tab:subtask-main3}). However, it lacks dynamic memory management: the KV cache grows without compression, eviction, or sliding window. This causes two issues in practice: GPU memory overflow as video length increases, and attention dilution that degrades generation quality in long sessions. We therefore reset the session at fixed intervals. RTA results indicate an optimal session length of about 1 minute (54.10\% overall), with poorer performance at both shorter (50.60\% at 1s) and longer intervals (30.15\% at 10min). At longer intervals, the model often produces incoherent or off-topic responses, confirming that naive KV cache accumulation without memory management is insufficient for sustained streaming. This highlights an open challenge: efficiently managing long-term memory under bounded computational resources remains unresolved for streaming MLLMs.

% —the KV cache grows unboundedly without any compression, eviction, or sliding window mechanism. This design leads to two practical issues: GPU memory overflow as video duration increases, and attention dilution that degrades generation quality over extended sessions. We therefore manually reset the session at fixed intervals. RTA results reveal an optimal session length around 1 minute (54.10\% overall), with performance declining at both shorter (50.60\% at 1s) and longer intervals (30.15\% at 10min). The degradation at longer sessions manifests as incoherent outputs and off-topic responses, confirming that naive KV cache accumulation without intelligent memory management is insufficient for sustained streaming operation. This finding underscores an open challenge: how to efficiently manage long-term memory under bounded computational resources remains an unsolved problem for streaming multimodal models.

\subsection{MPT in Current Models}
We proposed Memory Persistence Time (MPT) to measure how long a model retains information after initially answering correctly. However, the current evaluation does not include MPT results, as none of the tested models are suitable for this metric.
For pseudo-streaming models (VideoChat-Online, Qwen2.5-VL, Qwen2.5-Omni), each inference call is independent with no cross-call memory. In this setting, MPT is ill-defined: if a memoryless model answers correctly in both the initial and recall phases, we cannot tell whether it ``remembered'' the information or simply ``guessed correctly twice''.
For MiniCPM-o, although the KV cache can persist across calls, the lack of dynamic memory management forces us to reset sessions every few minutes. When an initial test and its recall test span a session reset, the model's memory is cleared in between. Any recall failure may then result from genuine forgetting or from the reset itself, making MPT measurements unreliable.

MPT is therefore a forward-looking metric that aligns with TeleEgo's goal of evaluating realistic long-term egocentric assistants. At present, models with true long-horizon streaming memory are still rare: few can run continuously for hours without resets while maintaining stable generation quality. As stronger streaming MLLMs emerge, MPT will become central to measuring how long information is retained beyond the moment of first use. TeleEgo, together with our released MPT evaluation framework and code, provides a unified setting to assess both real-time correctness and long-term memory persistence, paving the way for future work on practical egocentric AI assistants.

\section{Conclusion}
% We present TeleEgo, an online, omni-modal, first-person benchmark grounded in real-world use. It is built from continuous, multi-participant, multi-day recordings that align egocentric video, ambient speech/dialogue, and textual timelines under a unified global clock. A contract-based annotation scheme binds each query to its required modalities and to precise, time-stamped evidence spans, enabling auditable attribution.
% %
% Centered on three capability axes—Memory, Understanding, and Cross-Memory Reasoning—TeleEgo offers a fine-grained task suite together with a strict streaming-only evaluation protocol. Models consume inputs once, and a response receives credit only if it arrives within a task-specific decision window and satisfies evidence-compliance constraints.
% %
% We employ Real-Time Accuracy as the main metric in our experiments, capturing both correctness and timing under online decision constraints. In addition, we introduce Memory Persistence Time as a forward-looking protocol for probing long-horizon retention in continuous streams, and release an accompanying evaluation framework for future systems with stronger long-term streaming memory, along with failure breakdowns over retention, retrieval, alignment, and timing.
% %
% We envision TeleEgo as an ecologically valid, diagnostically informative, and reproducible foundation for developing first-person assistants that must remember, align, and act in real time.
We present TeleEgo, an online, omni-modal, first-person benchmark grounded in real-world use. It is built from continuous, multi-participant, multi-day recordings that align egocentric video, ambient speech/dialogue, and textual timelines under a unified global clock. A contract-based annotation scheme binds each query to its required modalities and to precise, time-stamped evidence spans, enabling auditable attribution.
Centered on three capability axes—Memory, Understanding, and Cross-Memory Reasoning—TeleEgo employs a strict streaming-only evaluation protocol where models consume inputs once and responses must arrive within task-specific decision windows.
We employ Real-Time Accuracy as the main metric in our experiments, capturing both correctness and timing under online decision constraints. In addition, we introduce Memory Persistence Time as a forward-looking protocol for probing long-horizon retention in continuous streams, and release an accompanying evaluation framework for future systems with stronger long-term streaming memory, along with failure breakdowns over retention, retrieval, alignment, and timing.
We envision TeleEgo as an ecologically valid foundation for developing first-person assistants that must remember, align, and act in real time.

% In summary, we present TeleEgo, an online, omni-modal, first-person benchmark grounded in real-world use, built from continuous, multi-participant, multi-scene, multi-day recordings that align video, ambient speech/dialogue, and dual textual timelines under a unified clock, and equipped with a contract-based annotation scheme that binds each query to its required modalities and precise, time-stamped evidence spans for auditable attribution. Centered on three capability axes, Memory, Understanding, and Cross-Memory Reasoning, TeleEgo offers a fine-grained task suite and a strict streaming-only evaluation protocol: responses receive credit only if they arrive within task-specific decision windows and satisfy evidence compliance. Two complementary metrics, Real-time Accuracy, and Memory Persistence Time, jointly assess correctness, response timing, and long-horizon memory, while failure cases are decomposed into retention, retrieval, alignment, and timing to yield actionable diagnostics.  We envision TeleEgo as an ecologically valid, diagnostically informative, and reproducible foundation for building first-person assistants that must remember, align, and act in real time.

\paragraph{Acknowledgments.} 
We thank the AI Glasses product team at China Telecom for their enthusiastic support and valuable assistance. 
We are grateful to Professor Haiwei Wu’s research group at the University of Electronic Science and Technology of China for their great help in data annotation, validation, and the construction of QA pairs.

{
    \small
    \bibliographystyle{ieeenat_fullname}
    \bibliography{main}
}

% WARNING: do not forget to delete the supplementary pages from your submission 
% \input{sec/X_suppl}

\end{document}